%% file: acl2019.tex
\documentclass[11pt,a4paper]{article}
\usepackage[hyperref]{acl2019}
\usepackage{times,latexsym,url}

\usepackage{hyperref,soulutf8,amsmath,mathtools,booktabs,tabularx,float,multirow,stfloats}
\usepackage[english, plain]{fancyref}
\usepackage[english]{babel}
\usepackage[autostyle=true]{csquotes}
\usepackage{censor}
\usepackage[normalem]{ulem}

\input{material_colors.tex}
\input{colors.tex}
\input{tikz.tex}
\usepackage{graphicx}

\newif\ifWIP
\WIPtrue % Comment to remove fixme highlights

\newif\ifBlind
\DeclareMathOperator{\similar}{sim}
\DeclareMathOperator*{\argmax}{arg\,max}
\aclfinalcopy % Uncomment this line for the final submission
\title{SenseFitting: Sense Level Semantic Specialization of Word Embeddings for Word Sense Disambiguation}
\author{Manuel Stoeckel, Sajawel Ahmed, Alexander Mehler\\
  \textit{Text Technology Lab}\\
  \textit{Goethe University Frankfurt}\\
  Frankfurt, Germany \\
  \{manuel.stoeckel\}@stud.uni-frankfurt.de \\
  \{sahmed, mehler\}@em.uni-frankfurt.de \\}
\date{}

\begin{document}
\maketitle
\begin{abstract}
	We introduce a neural network-based system of \textit{Word Sense Disambiguation} (WSD) for German that is based on \textit{SenseFitting}, a novel method for optimizing WSD. 
	We outperform knowledge-based WSD methods by up to $25\%$ F1-score and produce a new state-of-the-art on the German sense-annotated dataset \textit{WebCAGe}. 
	Our method uses three feature vectors consisting of a) sense, b) gloss, and c) relational vectors to represent  target senses and to compare them with the vector centroids of sample contexts. 
	Utilizing widely available word embeddings and lexical resources, we are able to compensate for the lower resource availability of German.
% 	In order to compensate for limited resources (as in the case of German), we rely on the optimization of raw text and lexical resources. % TODO DONE Review #1 & #2: consider rephrasing. "Utilize widely available resources.."
	%
	\textit{SenseFitting} builds upon the recently introduced semantic specialization procedure \textit{Attract-Repel}, and leverages sense level semantic constraints from lexical-semantic networks (e.g.\  GermaNet) or online social dictionaries (e.g.\ Wiktionary) to produce high-quality sense embeddings from pre-trained word embeddings. 
	% TODO DONE Review #2: fix sense level -> sense level
	%We evaluate our sense embeddings with a new sense-based similarity dataset built from SimLex-999 which we annotated for the sake of this work. 
	We evaluate our sense embeddings with a new SimLex-999 based similarity dataset, called \textit{SimSense}, that we developed for this work.
	We achieve results that outperform current lemma-based specialization methods for German, making them comparable to results achieved for English.
	% TODO: SenseFitting als zentraler Punkt/Grund für bessere performance hervorheben
\end{abstract}

\section{Introduction}
	Embeddings have been used successfully for a variety of Natural Language Processing (NLP) tasks, including Word Sense Disambiguation (WSD).
	There is a multitude of English evaluation datasets, sense annotated corpora, and tasks available to measure the performance of WSD systems, some of which were designed with embeddings in mind \citep{Moro:Navigli:2015}, some without \citep{Edmonds:Cotton:2001,Mihalcea:Chklovski:2004}.
	For German, however, only a few sense annotated datasets exist and therefore possibilities for NLP applications are limited.
	
	In this paper, we present a system for \textit{lexical-sample} WSD that is applicable to high and low resource languages by using widely available resources such as word embeddings trained on large unlabeled text corpora, and lexical-semantic networks.
	We refine the procedure of \textit{Simple Embedding-Based Word Sense Disambiguation} introduced by \citet{Oele:vanNoord:2018} for Dutch (another low-resource language), who use sense embeddings and expanded glosses to represent target senses. % TODO: rephrase low-resource language terminus?
	Instead of only extending glosses and contexts, as suggested by the authors, we include a third feature vector in sense representation that is derived from the sense relationships of the target senses in a lexical-semantic network.
	
	The sense embeddings used in this work are computed with the help of \textit{SenseFitting}, a novel method that extends the semantic specialization procedure Attract-Repel \citep{Mrksic:2017} by exploring sense level semantic constraints. % instead of lemma level ones.
	We apply SenseFitting to sense embeddings based on pre-trained word embeddings using a technique similar to that of \citet{Chen:Liu:2014}.
	We show that this method can significantly improve the performance of WSD compared to the use of word embeddings alone. 
	We also show that the quality of semantic constraints plays a major role in this improvement.
	
	We independently leverage both GermaNet \citep{Hamp:Feldweg:1997}, a German lexical-semantic network similar to WordNet \citep{Kunze:Lemnitzer:2002}, and the German Wiktionary\footnote{\url{https://de.wiktionary.org}}, a collaborative online social dictionary, to extract semantic constraints on the sense level and apply SenseFitting to word embeddings trained on two large German corpora (Leipzig \citep{Goldhahn:Eckart:2012} and COW \citep{Schafer:2015}).
    To evaluate our sense embeddings, we create \textit{SimSense}, a resource for German based on from SimLex-999 \citep{Hill:Reichart:2014} that addresses sense level semantic similarities.

	From the few open source datasets containing German sense annotations, we chose WebCAGe \citep{Henrich:Hinrichs:2012}, which is the only dataset that contains sense annotations from both lexical resources used in this paper: %, GermaNet and Wiktionary.
	WebCAGe mainly collects sentences sampled from the German Wiktionary and annotated with senses from GermaNet.
	We align WebCAGe with a current version of the German Wiktionary to use the resulting subset \textit{WebCAGe-aligned} (WCA) for evaluating our algorithm by means of both GermaNet and Wiktionary.
    
	The remainder of this paper is organized as follows: \Fref{sec:related-work} reviews related work. \Fref{sec:methods} describes SenseFitting and our WSD algorithm. \Fref{sec:setup} describes the preparation of the resources (GermaNet, Wiktionary, SimSense, WCA) used to evaluate our algorithm. \Fref{sec:results} presents experimental results. Finally, \fref{sec:conclusion} draws a conclusion.

\section{Related Work}
\label{sec:related-work}
    Recent years have seen multiple noteworthy approaches to WSD using sense embeddings.
    Early works trained sense embeddings on automatically disambiguated corpora.
    These include \citet{Chen:Liu:2014}, who present a unified model for joint word sense representation and disambiguation, and \citet{Iacobacci:Pilehvar:Navigli:2015}, who train sense embeddings using word2vec \citep{Mikolv:Chen:2013}.
    More recently, \citet{Uslu:et:al:2018} used a supervised method to train sense embeddings on manually annotated corpora.
    
    \citet{Oele:vanNoord:2018} use sense embeddings created with AutoExtend \citep{Rothe:Schutze:2015},
    a method for fine-tuning word embeddings to include representations of lexemes and synsets from WordNet. % in a single vector space.
    Tuning methods are also proposed by \citet{Faruqui:Dodge:2014}, who tune embeddings using synonymy-related constraints on lemma-level extracted from WordNet, and \citet{Wieting:Bansal:2015}, who explore paraphrase-related constraints.
    \citet{Mrksic:2016,Mrksic:2017} evaluate antonymy- and synonymy-related constraints on lemma-level to refine word embeddings, a method called \textit{semantic specialization}.
    
    In this article we combine the method of semantic specialization with the generation of sense embeddings. That is, we examine sense-related constraints provided by known lexical resources for the fine tuning of sense embeddings to finally solve various tasks of WSD.

\section{Methods}
\label{sec:methods}

    \begin{figure*}[ht!]
        \centering
        \includegraphics[width=\linewidth]{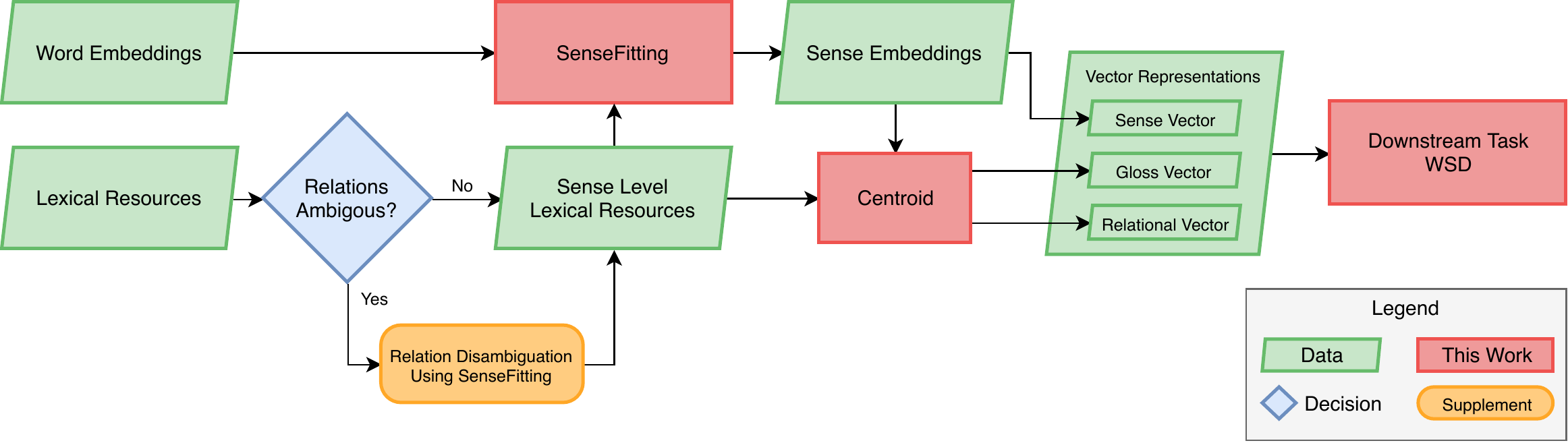}
        \caption{Flowchart showing the data flow within our work. Various data resources (green) are used in the presented methods (red) to improve the final WSD performance.} % TODO
        \label{fig:SenseFitting-flowchart}
    \end{figure*}
    
    \Fref{fig:SenseFitting-flowchart} shows a flowchart of our approach.
    Pre-trained word embeddings and sense level relational data are used in \textit{SenseFitting} to create sense embeddings.
    These are used along the lexical resources to create three-fold vector representations which are finally applied to our WSD downstream task.
    The following \fref{sec:three-fold vector representations} describes our vector representations of senses, while \fref{sec:sense-fitting} describes \textit{SenseFitting} and \fref{sec:wsd} describes our WSD approach.
    Some lexical resources, such as Wiktionary, are ambiguous on the sense level and have to be disambiguated first.
    This is briefly reported in \fref{sec:setup-sense-fitting-pre-wkt} and discussed more thoroughly in \fref{sec:supplement-sense-fitting-pre-wkt}.
    
    % The flowchart of our approach is shown in \fref{fig:SenseFitting-flowchart}: Starting with pre-trained word embeddings and sense level relational data, SenseFitting is used to generate sense embeddings that are finally used in downstream WSD tasks.
    % %
    % SenseFitting is described in \fref{sec:sense-fitting}, and our algorithm for WSD used in downstream-tasks in \fref{sec:wsd}.
    % %
    % Both of them relate to our three-fold vector representations of senses described in \fref{sec:three-fold vector representations}.
    % %
    % One of our chosen lexical resources uses ambigous relations, which have to be disambiguated first.
    % %
    % This process is described in \fref{sec:supplement-sense-fitting-pre-wkt}

\subsection{Three-fold Vector Representations of Senses} % TODO: Mir gefällt dieser Titel zwar, aber ggf. findet sich noch etwas knackigeres?
    \label{sec:three-fold vector representations}
		We use three feature vectors to represent any sense $S$ of a word $w$: 
		the sense vector $S_s$, the gloss vector $S_g$ and the sense relation vector $S_r$.
		These three vectors are compared with lexical contexts $C$ in which $w$ is observed to compute similarity scores between $C$ and $S$.
		Contexts $C$ in which $w$ has to be disambiguated are represented by the centroid $C_w$ of the embeddings of all textual neighbors of $w$ in $C$. % TODO DONE: not the lexical neighbors, the textual neighbors.
		Likewise, the gloss vector $S_g$ corresponds to the centroid of the embeddings of all words describing $S$ in the underlying gloss, while the sense relation vector $S_r$ denotes the centroid of the lexical neighborhood of $S$ in GermaNet and alternatively in Wiktionary.
		%
% 		Thirdly, the sense vector $S_s$ is obtained by post-processing $S_g$ using SenseFitting, a semantic specialization method that explores sense level semantic constraints to refine word embeddings (see \fref{sec:sense-fitting}). % replaced in favor of:
 		Thirdly, the sense vector $S_s$ is obtained from a sense embedding which has been created using SenseFitting, a semantic specialization method that explores sense level semantic constraints to refine word embeddings (see \fref{sec:sense-fitting}). %.. which suites the top-down approach with a possible "standalone" SenseFitting better
		
		To compute the relational vector $S_r$ we exploit explicit semantic constraints from terminological ontologies.
		This is inspired by \citet{Oele:vanNoord:2018}, who use elements of lexical chains that reach a similarity threshold or are similar according to a distribution thesaurus to extend the gloss of a sense.
		However, we select words or senses that are related to sense $S$ according to the given lexical-semantic network to form $S_r$ as an \textit{additional feature vector}.
		We assume that words contained in the gloss of a sense are distributed differently than words that have an explicit semantic relation to that sense.
		Thus, we distinguish $S_g$ and $S_r$ as two different vector representations.
		In order to compute $S_r$, we exploit synonymy and hyponymy relations from GermaNet and alternatively from Wiktionary.
% 		\textcolor{red}{In order to compute $S_r$ we exploit WHICH sense relations from GermaNet and WHICH sense relations from Wiktionary.}

	\subsection{SenseFitting}
	\label{sec:sense-fitting}
		Sense vectors $S_s$ %in \Fref{eq:1} 
		are computed by means of \textit{SenseFitting}, an %\comment{red800}{novel} % TODO DONE: removed "novel"
		optimization method of semantic specialization that creates sense embeddings from word embeddings using lexical-semantic networks.
		SenseFitting is based on \textit{semantic specialization}, a post-processing method that increases the quality of embeddings by constraining their computation through semantic constraints derived from terminological ontologies.
		In previous work, these constraints are extracted on the \textit{lemma level} to specialize \textit{word} embeddings \citep[e.g.][]{Faruqui:Dodge:2014, Mrksic:2016, Mrksic:2017}.
		We extend this approach by constraining the computation of \textit{sense} embeddings through \textit{sense level} relations. % constraints. % TODO: fix double constrain
	
		Initially, we compute the \textit{sense} embedding $S_s % \gets S_g
		$ of the sense $S$ of word $w$ as the centroid of the embeddings of all words contained in the gloss describing $S$ whose cosine similarity to the \textit{word} embedding of $w$ is higher than threshold $\delta$ (we set $\delta = 0.05$), similarly to the way \citet{Chen:Liu:2014} create their initial sense embeddings.
		To keep distributional information from the original embeddings and to compensate for differences in the quality of glosses, we include the embedding of $w$ into computing this centroid.
		In a second step, $S_s$ is post-processed with the help of semantic specialization.
		This is done using the Attract-Repel\footnote{\url{https://github.com/nmrksic/attract-repel}} algorithm \citep{Mrksic:2017}.
		This specialization algorithm differs from related approaches such as retrofitting \citep{Faruqui:Dodge:2014} in that it refers to synonyms and antonyms to constrain the fine-tuning of embeddings in different ways:
		while training attempts to locate the embeddings further apart from antonyms, synonyms are brought closer together. 
		In this way, embeddings are created that better reflect sense relations.
		As a result of this constraint satisfaction process, $S_s$ is computed as a hybrid vector representation to which $S_g$ and the sense relations that embed $S$ into the operational lexical resource (GermaNet or Wiktionary) contribute in different ways. % TODO: technically S_g does not contribute per se, as it contains *all* words in the gloss, while the sense vector initialization only contains words which satisfy the similarity contraint delta. While these two vectors might be identical, they don't have to be, eg. the gloss contains antonyms which are already dissimilar "ABC [1]: the opposite of XYZ". 
	
\subsection{Word Sense Disambiguation}
    \label{sec:wsd}

		After $S_g, S_r$ and $S_s$ have been calculated for each sense $S$ of a word $w$, WSD can be performed.
		%
		%The WSD algorithm creates a representation for each candidate sense $S$ of the target word and computes its similarity with the context vector $C$.
		For this purpose, we calculate the similarity of the context vector $C_w$ and the three-part vector representation $S_w = \{S_s, S_g, S_r\}$ of the sense $S$ of $w$, choosing the sense with the highest score as follows:
		\begin{equation}\label{eq:1}
            \similar(C_w, S_w) = \dfrac{1}{|S_w|} \sum\limits_{X~\in~S_w} \cos(C_w,X)
		\end{equation}
% 		%\textcolor{red}{[Hier hätte es nahegelegen, das nicht gleichgewichtet additiv zu machen, sondern gewichtet und die Gewichte der Komponentenvektoren zu lernen!]}
% 		%
% 		The sense with the highest score is chosen for disambiguation:
% 		%
		\begin{equation}\label{eq:2}
		    S_{\mathit{opt}} = \argmax\limits_{S_w ~\in~ S(w)} \similar(C_w, S_w)
		\end{equation}
		Here $S(w)$ denotes the set of all senses of $w$.
	    Note that all three component vectors are equally weighted in \Fref{eq:1}. 
	    Of course, we could have considered weighting the effects of these component vectors differently and learning their relative weights. 
	    In our experiments, however, our simpler approach was already very effective.

\section{Experimental Setup}
\label{sec:setup}

	\subsection{Embeddings}
	\label{sec:setup-embeddings}
	    % TODO Review #1: licence & availability
		We used a set of pre-trained word embeddings\footnote{\url{https://texttechnologylab.org/resources2018/}} of \citet{Ahmed:Mehler:2018} as a basis for SenseFitting.
		This concerns \textit{wang2vec} embeddings \citep{Ling:Dyer:2015} trained on two large German corpora: 
		the \textit{COW} corpus and the \textit{LeipzigMT} corpus, which is the combination of \textit{Leipzig40-2018} \citep{Schafer:2015} and WMT-2010-German corpora \citep{Callison:2010}.
		\citet{Ahmed:Mehler:2018} trained Structured Skip-gram wang2vec embeddings with a dimension of 100, window size of 8, minimum word count threshold of 4 and otherwise default parameters as given in \citet{Ling:Dyer:2015}.
		
		As German is a highly inflected language in comparison to English, a single word can be distributed across various morphological and spelling variants effectively weakening the information value of their embeddings.
		To mitigate the dispersion of information of an embedding over multiple variants, we also use embeddings trained on lemmatized versions of the corpora. % TODO Review #2: Details about the embeddings, effect of lemmatization & lowercasing
		
		The original embeddings are very large, based on a vocabulary of several million words. 
		Thus, we drop all entries that do not occur anywhere in Wiktionary, GermaNet or our test data.
		In the case of COW, this leads to $435\,003$ embeddings of words converted to lowercase letters and $402\,575$ lemma-based embeddings.
		In the case of Leipzig40MT, this leads to $347\,550$ lowercased word embeddings and $334\,625$ lemma embeddings. %, respectively.
		On top of these embeddings, we performed SenseFitting using the default parameters for Attract-Repel defined by \citet{Mrksic:2017}.
		
	\subsection{Preparing SenseFitting}
	\label{sec:sense-fitting-pre}
	    We utilize GermaNet and Wiktionary as sense inventories for SenseFitting.
	    Sections \ref{sec:setup-sense-fitting-pre-gn} and \ref{sec:setup-sense-fitting-pre-wkt} describe the steps necessary to prepare them for SenseFitting, while \fref{sec:setup-sf-resources} provides a summary of the resulting resources.
	    \Fref{sec:setup-sf-SimSense} describes SimSense, a sense-annotated semantic similarity evaluation dataset, which we specifically created for this work.
	    
    	\subsubsection{Preprocessing GermaNet}
    	\label{sec:setup-sense-fitting-pre-gn}
    		Only a few modifications are necessary to prepare GermaNet for SenseFitting with Attract-Repel.
    		GermaNet does not contain own sense glosses but glosses mapped from Wiktionary by \citet{Henrich:Hinrichs:2011}, yielding $27\,903$ sense definitions.
    		We expand these definitions with an additional $32\,850$ mappings from \citet{Matuschek:Gurevych:2013}.
    		In this way, we obtain a total of $30\,352$ distinct sense glosses for our training\footnote{The entire joined dataset holds $30\,455$ mappings, but some do not occur in our GermaNet version or are just empty.}.
    	    In order to run SenseFitting on the lemma-based wang2vec embeddings from \citet{Ahmed:Mehler:2018}, we lemmatized the glosses using spaCy V2.0\footnote{\url{https://spacy.io}}, the same tool used to lemmatize the embeddings' training corpora.
    	
    	\subsubsection{Preprocessing Wiktionary}
    	\label{sec:setup-sense-fitting-pre-wkt}
    		
    		\begin{figure}[t]
    		    \centering
    		    \input{wiktionary_example.tikz}
    		    \caption{An example Wiktionary sense-to-lemma relation in solid green and a disambiguated relation as dashed line below. Image courtesy of \citet{Mehler:Gleim:2018}} % TODO: caption & citation
    		    \label{fig:wiktionary-example}
    		\end{figure}
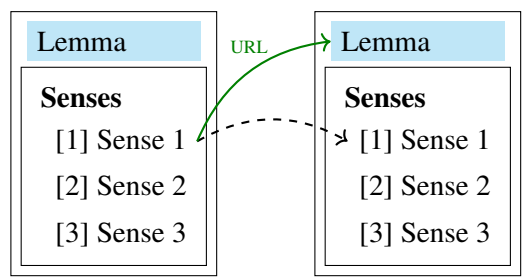
    	    
    	    We used WikiDragon \citep{Gleim:Mehler:Song:2018} to obtain a Neo4j\footnote{\url{https://neo4j.com/}} graph-database from a current German Wiktionary dump (April 2018) and used spaCy V2.0 to lemmatize $147\,363$ sense glosses, adding them to the database.
            Utilizing the German Wiktionary for SenseFitting, which relies on sense-to-sense relations, is more challenging because relations are encoded as URLs from the source sense to the target's page (sense-to-lemma relations, see \fref{fig:wiktionary-example}) and thus are inherently ambiguous.
            In a further pre-processing step, detailed and evaluated in \fref{sec:supplement-sense-fitting-pre-wkt}, we are able to disambiguate $164\,154$ of $220\,927$ relations\footnote{Counting only antonyms, hyponyms, hypernyms and synonyms.}, leaving $56\,773$ ambiguous sense-to-lemma relations.
            These are still used in training, as our experiments have shown a small positive effect on embedding quality while retaining them.
    		
		\subsubsection{SenseFitting Training Resources}
		\label{sec:setup-sf-resources}
		
    		After pre-processing, the relation count in the case of GermaNet ranges from $685\,484$ synonym relations to $3\,426$ antonym relations.
    		In the case of Wiktionary, the relation count ranges from $249\,751$ synonyms to $51\,173$ antonyms%
    		\footnote{These are the numbers when using the COW embeddings.
    		With the Leipzig embeddings, all numbers are about 10-20\% lower.}.
    		\Fref{tab:sense-inv-stats} in the supplemental material \fref{sec:supplement-sensefitting-training-resources} shows all resource statistics including the individual count for each relation type.
    		%
    % 		Wiktionary relations that could not be disambiguated are used as sense-to-lemma relations.
    		%
    % 		Rows marked with $^*$ show the number of annotated relations, the ones marked with $^+$ show the total number of relations used. TODO: move these rows?
    		%
    		Training with this amount of data takes about 15 minutes per epoch for Wiktionary and 30 minutes for GermaNet on a single NVIDIA 1070 GPU.

		\subsection{SenseFitting Evaluation Dataset}
		\label{sec:setup-sf-SimSense}

		\begin{table*}[t!]
			\centering
			\small
			\begin{tabularx}{\textwidth}{lllXllc}
				\toprule
				Dataset & Sample & Source & Source Gloss & Target & Target Gloss & Score \\
				\midrule
				SimLex & 1 & new & N/A & fresh & N/A & 8.62 \\
				SimLex & 2 & wide & N/A & fresh & N/A & 0.00 \\
				\midrule
				SimSense & 1 (pos) & new$_1$ & Recently made, or created. & fresh$_1$ & Newly produced or obtained. & 9.00 \\
				SimSense & 2 (neg) & new$_1$ & Recently made, or created. & fresh$_2$ & Not cooked, dried, frozen, or spoiled. & 3.00 \\
				SimSense & 3 (neg)& new$_1$ & Recently made, or created. & fresh$_4$ & Refreshing or cool. & 0.00 \\
				SimSense & 4 (false) & wide$_1$ & Having a large physical extent from side to side. & fresh$_1$ & Newly produced or obtained. & 0.00 \\
				\bottomrule
			\end{tabularx}
			\normalsize
			\caption{\label{tab:simsense-example} SimSense annotations using English Wiktionary glosses for two word pairs from the English SimLex. Index $n$ denotes the $n$-th sense in the Wiktionary entry.}
		\end{table*}
		
			In their work, \citet{Mrksic:2017}\ measure Attract-Repel's performance by using the multilingual semantic similarity evaluation dataset SimLex \citep{Hill:Reichart:2014}.
			It contains word pairs that have been annotated by experts regarding their degree of similarity in order to clarify their status as synonyms.
			\citet{Mrksic:2017} obtain state-of-the-art results using Spearman's rank correlation coefficient to measure the performance of their embeddings.
			Although SimLex is available in German, there is, unfortunately, no such dataset for sense similarity.

			Thus, in order to evaluate our sense embeddings computed with SenseFitting, we tagged a subset of SimLex word pairs with senses using the GermaNet sense inventory and annotated their degree of similarity.
			
    		\subsubsection{Dataset Description} % TODO: section title
    		\label{sec:}
    			The annotations are grouped into three categories: positive, negative and false samples.
    			They aim to capture four different types of relations between senses for which we believe a semantic specialization method should be tested.
    			\Fref{tab:simsense-example} exemplifies annotations for each relation type using two English word pairs from SimLex\footnote{In this example, we consider \textit{new$_1$} and \textit{fresh$_1$} to be synonyms.} using sense glosses from the English Wiktionary.

    			\textit{Positive samples} are obtained from word pairs that are connected at least by one sense relation that exists between their senses.
    			We look for such pairings for each sense in our subset of SimLex using GermaNet as resource. 
    			From the first pairing that we find for such a sense, we create a \textit{positive sample}.
    			Sample \textit{1 (pos)} in \fref{tab:simsense-example} captures two synonymous senses of \textit{new} and \textit{fresh}.

    			Then, for any such pairing, we check all other pairings and use the first one whose senses are unrelated to generate a \textit{negative sample}.
    			Sample \textit{2 (neg)} shows a negative sample of two unrelated senses with a small similarity, while sample \textit{3 (neg)} shows a negative sample of two unrelated senses with no similarity.

    			Words and senses for which we did not find any positive sample are considered to be entirely unrelated, like sample \textit{4 (false)} in the table above. 
    			In this case, we use the first senses of such words to generate \textit{false samples}.
    			Using synonym, antonym, hyponym and hypernym relations from GermaNet for generating positive samples, we get 210 pairs, which were finally annotated regarding their similarity based on their glosses (see \fref{sec:setup-sf-SimSense-annotation}).
    			These annotations form the evaluation dataset called \textbf{SimSense}.

    			Unfortunately, GermaNet does not feature Wiktionary identifiers for its glosses mapped by \citet{Henrich:Hinrichs:2011}, complicating the use of SimSense to evaluate SenseFitting performance with Wiktionary.
    			Thus, only 76 of the annotated pairs are applicable to evaluate SenseFitting with Wiktionary.
		
    		\subsubsection{Annotation Process}
    		\label{sec:setup-sf-SimSense-annotation}
    		    The SimSense dataset was annotated by two native speakers of German.
    		    Both annotators were given a table containing a single unannotated SimSense pair per row.
    		    Each row only detailed the senses lemmata and definition, but did not contain further information like the sample type.
    		    With the practical example in \fref{tab:simsense-example}, the annotators would have been given the third to sixth column (Source, Target and respective Gloss columns).

    		    To evaluate the resulting dataset, we computed inter-annotator agreement $\rho$ and average response deviation $\sigma$ in the same manner as \citet{Hill:Reichart:2014}.
    		    The authors compute $\rho$ as the average of the Spearman rank correlation for each annotation pairwise across the authors and $\sigma$ as the average standard deviation, likewise. % TODO: formulation

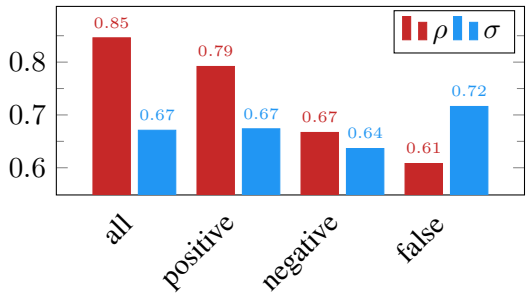
\begin{figure}[ht!]
                    \centering
    			    \input{inter-annotator_agreement.tikz}
    			 %   \vspace{-5pt}
                    \caption{Inter-Annotator Agreement $\rho$ \& Response Consistency $\sigma$ for SimSense}
                    \label{fig:SimSense-IAA}
                \end{figure}

    		    \Fref{fig:SimSense-IAA} shows the inter-annotator agreement and average response deviation for all pairs and by sample type.
    		    With $0.85$ inter-annotator agreement $\rho$ over all annotation is significantly ($+0.177$) higher than for SimLex-999.
    		    Further, the response deviation $\sigma$ is about half compared with SimLex-999 \citep{Hill:Reichart:2014}\footnote{Note: The authors report a \textit{response consistency} of $0.751$ for all annotations in their work, which is defined as $1/\sigma$.}.
    		    We presume the lower ambiguity of senses with a given definition to be responsible for our better values, possibly influenced by the lower amount of annotators and annotation pairs.
			
	\subsection{Word Sense Disambiguation}
		In all experiments described below, our WSD method is based on the following parameter setting unless otherwise stated.
		We set the window size to observe context words around target words to 8, matching the window size of the wang2vec embeddings during training.
		We exclude the target word from the list of context words, as suggested by \citet{Iacobacci:Pilehvar:2016}.
		As a measure of the performance of WSD, we adopt the scoring system used in IMS \citep{Zhong:Ng:2010}, taking a single best guess for every sample.
		
		\subsubsection{Preprocessing WebCAGe}%Alignment
		\label{sec:setup-wca}
			
			As mentioned at the beginning, evaluation datasets such as Senseval or SemEval, as well as training datasets of sense annotations, are missing in German.
			However, there is a single sense annotated corpus that has annotations for GermaNet and Wiktionary called \textit{WebCAGe} \citep{Henrich:Hinrichs:2012}.
			WebCAGe is a corpus mainly composed of examples from word sense definitions from the German Wiktionary which were automatically annotated with GermaNet senses and hand-corrected afterward.
			However, the authors do not provide a mapping to Wiktionary; only GermaNet senses are tagged with an ID.
			After the pre-processing outlined in the supplemental section \fref{sec:supplement-wca}, we get a new corpus called \textit{WebCAGe-aligned} (WCA) that contains about 58\% of WebCAGe's sentences, for a total of 4\,125 sentences with 2\,178 unique lemmas and thus 1.89 sentences per lemma.
			
\section{Results}
\label{sec:results}
% 	\subsection{Intrinsic Evaluation of SenseFitting}
    \subsection{SenseFitting Performance}
		\label{sec:results-sf}
			\Fref{tab:SimSense-scores} shows the Spearman's rank correlation for all our created sense embeddings with SimSense.
			The third column shows the specialization method used for the given embedding.
			Here \enquote{Baseline} denotes sense embeddings which were initialized using only the vector centroid of sense glosses as described in \fref{sec:sense-fitting} but not specialized further.

			All results for GermaNet outperform the baseline by about $+0.314$, across both COW and Leipzig embeddings and with or without lemmatization.
			This shows that extending word embeddings to the sense-layer and specializing them using sense-to-sense relations improves the representation of semantic similarity by a large margin.
			For comparison to lemma-based specialization, we ran Attract-Repel with GermaNet lemma relations on our COW embeddings.
			The resulting embeddings scored $0.415$ for SimLex, slightly lower than the ones trained by \citet{Mrksic:2017} which scored $0.43$.
			The underlying COW embeddings however already score $0.352$ for Simlex, resulting in a performance increase with Attract-Repel of $0.063$.
			Therefore, the increase of SenseFitting is nearly 7 times as large as the increase of Attract-Repel using lemma-to-lemma constraints.
			This improvement is also significantly higher than the improvement with Attract-Repel on SimLex for German reported by \citet{Mrksic:2017} of $+0.23$ using lemma-based monolingual data. %, and the improvement using cross-lingual data of $+0.34$.
			
			\begin{table}[hbt!]
				\centering
				\small
				\begin{tabularx}{0.95\linewidth}{Xllc}
					\toprule
					Embedding & Resource & Method & $\rho$ \\
					\midrule
					Leipzig-lower & GN & Baseline & $0.385$ \\
					Leipzig-lower & GN & SenseFitting & $0.700$ \\
					\midrule
					Leipzig-lemma & GN & Baseline & $0.359$ \\
					Leipzig-lemma & GN & SenseFitting & $0.703$ \\
					\midrule
					COW-lower & GN & Baseline & $0.425$ \\
					COW-lower & GN & SenseFitting & $\mathbf{0.738}$ \\
					\midrule
					COW-lemma & GN & Baseline & $0.420$ \\
					COW-lemma & GN & SenseFitting & $0.734$ \\
					\midrule\midrule
					Leipzig-lower & WKT & Baseline & $0.212$ \\ % $0.198$ \\
					Leipzig-lower & WKT & SenseFitting & $0.484$ \\ %$0.349$ \\
					\midrule
					Leipzig-lemma & WKT & Baseline & $0.236$ \\ % $0.118$ \\
					Leipzig-lemma & WKT & SenseFitting & $0.392$ \\ % $\mathbf{0.381}$ \\
					\midrule
					COW-lower & WKT & Baseline & $0.197$ \\ % $0.214$ \\
					COW-lower & WKT & SenseFitting & $\mathbf{0.488}$ \\ % $0.370$ \\
					\midrule
					COW-lemma & WKT & Baseline & $0.216$ \\ % $0.108$ \\
					COW-lemma & WKT & SenseFitting & $0.466$ \\ % $0.298$ \\
					\bottomrule
				\end{tabularx}
				\normalsize
				\caption{\label{tab:SimSense-scores} SimSense Spearman's correlation $\rho$ with COW and Leipzig embeddings. Bold results are the maxima for the corresponding resource.} % \vspace*{-10pt} 
			\end{table}
            
			It is worth noting that all COW embeddings trained with GermaNet relations strictly outperform the Leipzig embeddings confirming \citet{Ahmed:Mehler:2018} observation that the COW embeddings are of higher quality.
			The authors attribute this to the significantly higher data size of the COW corpus which is with over $600$ million sentences more than ten times larger than the extended Leipzig corpus with $60$ million sentences.
			
			The lower half of \Fref{tab:SimSense-scores} shows the Spearman's correlation for the Wiktionary sense embeddings, which perform only about half as good as GermaNet's did.
			Using GermaNet to create the SimSense dataset may introduce a bias, as we do not control whether Wiktionary has similar relations to the ones from which the dataset is induced.
			Instead we rely on the fact that the glosses in GermaNet were automatically mapped from Wiktionary \citep{Henrich:Hinrichs:2011}, strongly suggesting there must a certain degree of similarity to the sense definitions and relations.
			Still, further work has to be done to fully analyze the possible impact of this issue and refine the evaluation of SenseFitting for Wiktionary.
			While the total performance using Wiktionary as SenseFitting resource is lower when compared with GermaNet, the relative increase of $+0.291$ for the COW-lower embeddings using SenseFitting is still noteworthy.
			%
% 			Together with the results of Wiktionary relation disambiguation from \Fref{tab:SenseFitting-disambig} of 52\%-54\% F$_1$-score, our results indicate that the quality of the chosen constraint resource is of high significance when specializing embeddings and sense-to-lemma constraints fail to replace sense-to-sense ones.
			
    		Our results demonstrate that we can improve the semantic quality of embeddings even more with sense level constraints.
			Further, our results show that SenseFitting improves the representation of semantic relationships not only for senses with an explicit relationship, but also for those not covered in the lexical resource.
			Given the status of German as a low-resource language and the difficulties in generating embeddings for this highly inflectional language, these results are promising.

	\subsection{Word Sense Disambiguation}
	\label{sec:results-wsd}
	    Taking all optimized resources into account, we come to the main task of WSD.
		The following two tables show the results of our WSD method for the WCA dataset.
		\Fref{tab:wsd-wca-gn} holds the results with GermaNet as sense inventory, while \Fref{tab:wsd-wca-wkt} holds the results for Wiktionary:
		%
		%For our work (all but the first row) 
		w2v stands for wang2vec, AR for Attract-Repel and SF for SenseFitting.
		When running our disambiguation with word embeddings (w2v \& AR), we drop the sense feature vector and use gloss and relational features only.
	    We used a First Sense baseline, as suggested by \citet{Iacobacci:Pilehvar:2016}, and a Random Sense baseline.

	    All results for GermaNet outperform these baselines by a significant margin.
	    The embeddings trained on the Leipzig corpus are outperformed by the COW embeddings, again confirming the previous assumption that the latter are of higher quality.
		However, we cannot entirely confirm our hypothesis about lemmatization (formulated in \fref{sec:setup-embeddings}).
		The lemmatized Leipzig sense embeddings bring a slight improvement for verbs, beating their lowercased variant by $+0.98$, but the COW embeddings do not profit from lemmatization.
		In light of the fact that verbs usually have the most inflections in German, followed by adjectives, we presume that lemmatizing the many possible wordforms of a verb or adjective helps to capture their meaning and consequently improve their embedding.
		Although spaCy gives decent results for lemmatization in German, we might see an improvement with a better lemmatizer, as our disambiguation performance depends on it.	    
		\begin{table}[ht!]
			\centering
			\small
			\begin{tabularx}{\linewidth}{p{3em}Xccc}
				\toprule
				Model & Method & Nouns & Verbs & Adj. \\
				\midrule
				\multirow{2}{=}{\citet{Henrich:2015}} & Lesk & $53.17$ & $28.76$ & $29.01$ \\
				& Borda c. & $55.92$ & $45.97$ & $\mathbf{56.28}$ \\
				\midrule\midrule
				\multirow{2}{=}{Baseline} & First Sense & $38.68$ & $31.54$ & $33.04$ \\
				& Random Sense & $34.28$ & $26.59$ & $29.17$ \\
				\midrule\midrule
				w2v$^*$ & COW-lower & $51.49$ & $42.67$ & $50.89$ \\
				\midrule
				AR$^*$ & COW-lower & $46.57$ & $43.01$ & $46.43$ \\
				\midrule
				SF & Leipzig-lower & $50.00$ & $43.38$ & $49.70$ \\
				\midrule
				% SF & Leipzig-lemma & $49.64$ & $45.02$ & $52.55$ \\ % Using raw text with lemma embeddings
				SF & Leipzig-lemma & $46.88$ & $44.36$ & $47.46$ \\ % Using lemmatized text with lemma embeddings
				\midrule
				SF & COW-lower & \underline{$\mathbf{56.03}$} & \underline{$\mathbf{47.13}$} & \underline{$53.27$} \\
				\midrule
				% SF & COW-lemma & $54.05$ & $47.25$ & \underline{$54.82$} \\ % Using raw text with lemma embeddings
				SF & COW-lemma & $50.21$ & $45.68$ & $51.79$ \\ % Using lemmatized text with lemma embeddings
				\bottomrule 
			\end{tabularx}
			\normalsize
			\caption{\label{tab:wsd-wca-gn} WCA F$_1$-Scores for \textbf{GermaNet} in percent. Our maximas are underlined, absolute best in bold. $^*$Only definition and relational feature vectors are used.}  \vspace*{-4pt} % TODO: table spacing
		\end{table}
		
		We take steps towards closing the gap to WSD results for high-resource languages such as English, and outperform the Lesk-based methods used by \citet{Henrich:2015} on the original WebCAGe corpus with both GermaNet as dataset by $+2.86$, $+18.37$ and $+25.81$ for nouns, verbs, and adjectives, respectively.
		The work of \citet{Henrich:2015} provides the only other German WSD method for WebCAGe, using knowledge-based and supervised machine learning (ML) methods.
		The ML approaches are not comparable to our work, as, according to \citet{Henrich:2015}, they are only applicable to a very small subset of 43 out of 2\,178 lemmas from the WebCAGe corpus. % TODO DONE? Review #2: make more precise
		We outperform the author's best knowledge-based method \textit{Borda count}, which combines all methods in their work, for nouns ($+0.11$) and verbs ($+1.16$).
		Further, our sense embedding performance exceeds the Attract-Repel (AR) embeddings performance. 
		The AR embeddings, in turn, cannot beat the \textit{pure} wang2vec embeddings, although after specialization they perform similarly for SimLex ($0.415$) to the ones trained by \citet{Mrksic:2017} using German monolingual data ($0.43$).
	    
		\Fref{tab:wsd-wca-wkt} show the results using Wiktionary as a sense inventory.
		Wiktionary has a large number of sense glosses, resulting in good baseline performance using wang2vec embeddings, however, with automatically annotated relations, SenseFitting does not improve the performance over the original wang2vec embeddings consistently.
		Still, our results beat the Wiktionary Lesk results from \citet{Henrich:2015} by more than $5\%$. 
		The best results throughout our work are obtained with SenseFitting on expert-crafted lexical data from GermaNet, which outperforms Wiktionary in all tasks.
		\begin{table}[t!]
			\centering
			\small
			\begin{tabularx}{\linewidth}{p{3em}Xccc}
				\toprule
				Method & Model & Nouns & Verbs & Adj. \\
				\midrule
				\multirow{2}{=}{\citet{Henrich:2015}} & Lesk & $50.38$ & $41.66$ & $42.77$ \\
				& Borda c. & $55.92$ & $45.97$ & $\mathbf{56.28}$ \\
				\midrule\midrule
				\multirow{2}{=}{Baseline} & First Sense & $41.68$ & $38.06$ & $42.90$ \\
				& Random Sense & $29.97$ & $28.69$ & $32.63$ \\
				\midrule\midrule
				w2v$^*$ & COW-lower & \underline{$\mathbf{55.95}$} & $46.58$ & \underline{$50.76$} \\
				\midrule
				AR$^*$ & COW-lower & $53.96$ & $45.40$ & $49.55$ \\
				\midrule
				SF & Leipzig-lower & $47.68$ & $43.94$ & $46.43$ \\
				\midrule
				% SF & Leipzig-lemma & $47.88$ & $46.10$ & $45.92$ \\ % Using raw text with lemma embeddings
				SF & Leipzig-lemma & $45.77$ & $39.82$ & $46.67$ \\ % Using lemmatized text with lemma embeddings
				\midrule
				SF & COW-lower & $52.58$ & \underline{$\mathbf{47.04}$} & $48.81$ \\
				\midrule
				% SF & COW-lemma & $50.88$ & \underline{$\mathbf{48.73}$} & $50.25$ \\ % Using raw text with lemma embeddings
				SF & COW-lemma & $46.79$ & $40.91$ & $49.85$ \\ % Using lemmatized text with lemma embeddings
				\bottomrule 
			\end{tabularx}
			\normalsize
			\caption{\label{tab:wsd-wca-wkt} WCA F$_1$-Scores for \textbf{Wiktionary} in percent. Our maximas are underlined, absolute best in bold. $^*$Only definition and relational feature vectors are used.} \vspace*{-5pt} % TODO: table spacing
		\end{table}

\section{Conclusion}
\label{sec:conclusion}
    We introduced SenseFitting, an extension of semantic specialization algorithms based on Attract-Repel. 
    SenseFitting performs a semantic specialization of the centroids of word embeddings used to represent senses by means of sense level constraints.
    We evaluated our sense-specialized embeddings using a newly created resource called \textit{SimSense} and showed that SenseFitting improves sense representation by a large margin.
    Then, we applied our sense embeddings to Word Sense Disambiguation (WSD) using a simple disambiguation algorithm.
    Our sense embedding-based method outperforms competitors based on word embeddings, specialized word embeddings, the first sense baseline, as well as previous results obtained for the WebCAGe corpus.
    To sum up, our results show: a) SenseFitting generates embeddings that allow semantic similarities on the sense level to be reconstructed in a valid manner; 
    b) recognizing these similarities leads to an increase in performance in downstream-tasks such as WSD.
	
	Future work will examine whether the results obtained for German can be transferred to other languages.
	Further work will evaluate whether sense embeddings obtained by SenseFitting can increase performance while using other embedding based WSD methods.
	We believe that high-quality representations of senses and their similarities can enhance performance in tasks other than WSD.
	This will also be tested in future work.
	In addition, we plan to apply SenseFitting to pre-trained sense embeddings, such as FastSense \citep{Uslu:et:al:2018} to evaluate whether we can improve their performance as well.
	%, applying SenseFitting to English word embeddings using WordNet as a lexical resource and subsequently comparing our WSD method to state-of-the-art methods for the English language.
	%
% 	Finally, we will include ELMo embeddings \citep{Peters:2018} in our WSD algorithm since they model polysemy by processing words in context.
	
\bibliography{acl2019.bib}
\bibliographystyle{acl_natbib}

\appendix

\clearpage
\newpage
\section{Supplemental Material}
    \subsection{Prepraring SenseFitting}
    \label{sec:supplement-sense-fitting-pre}
    
        % \subsubsection{Preprocessing GermaNet}
        % \label{sec:supplement-sense-fitting-pre-gn}
        %     \textcolor{orange800}{TODO} % TODO
        
        \subsubsection{Preprocessing Wiktionary}
        \label{sec:supplement-sense-fitting-pre-wkt}
        
    	    We are able to disambiguate 164\,154 of 220\,927 relations, leaving 56\,773 ambiguous sense-to-lemma relations.
    	    Out of the disambiguated ones, there were $110\,584$ sense-to-lemma relations where the target lemma was monosemous, leaving $110\,343$ ambiguous relations of $220\,927$ relations in total.
    		To disambiguate the remaining $110\,3435$ ambiguous relations\footnote{Counting only antonyms, hyponyms, hypernyms and synonyms.}, we utilize SenseFitting itself.
    		We initialize a minimalist sense embedding by creating a copy of a lemma embedding for each of its senses and run SenseFitting for one epoch with sense-to-lemma relations, where a single epoch means a full run of Attract-Repel.
    		Using the resulting embedding, we try to disambiguate the target side of such a relation by selecting the sense with the lowest (or highest in the case of antonyms) cosine distance to the target sense.
    		We repeat this process with these partially disambiguated relations for up to 10 epochs, updating the relations each time.
    		During our experimentation, we noticed a significant performance decrease when disambiguating antonyms together with the other sense relations. 
    		Thus, we run the disambiguation in two batches: 
    		one for hyponyms, hypernyms and synonyms, and one for antonyms.
    		Finally, we can proceed with SenseFitting using the disambiguated relations.
    		%
    % 		The final number of sense glosses used in Wiktionary-related training is $132\,963$\footnote{The total amount is $147\,377$, but some glosses are empty as in the case of GermaNet.}.

		\subsubsection{Wiktionary Relation Disambiguation}
    		
			To evaluate the results of this automatic disambiguation, we used the sense-annotated Wiktionary created by \citet{Meyer:Gurevych:2012}.
			The authors created four datasets of different language combinations to evaluate the disambiguation performance of Wiktionary relations.
			The German dataset consists of $1\,119$ annotated relations of which $514$ are correct according to manual annotations. Of these, we are able to use $486$ for evaluation.

			To determine whether the performance of disambiguating relations during pre-processing is influenced by the structure of the used resource, we repeated the process described in \fref{sec:supplement-sense-fitting-pre-wkt} for GermaNet.
			We removed target sense information from all relations and ran SenseFitting for 10 epochs, which sums to a total of 50 Attract-Repel iterations.
			\begin{table}[ht!]
				\centering
				\small
				\begin{tabularx}{\linewidth}{Xlccc}
					\toprule
					Dataset & Relation type & P & R& F$_1$ \\
					\midrule
					\multirow{5}{=}{GermaNet} 
					& Synonyms  & 99    & 90    & 94 \\
					& Hypernyms & 94    & 85    & 90 \\
					& Hyponyms  & 96    & 90    & 93 \\
					& Antonyms  & 100   & 89    & 94 \\
					& All       & 96    & 88    & 92 \\
					\midrule
					\multirow{5}{=}{Wiktionary lower} 
					& Synonyms  & 71    & 34    & 46 \\
					& Hypernyms & 71    & 39    & 50 \\
					& Hyponyms  & 71    & 54    & 62 \\
					& Antonyms  & 70    & 46    & 56 \\
					& All       & 71    & 43    & 54 \\
					\midrule
					\multirow{5}{=}{Wiktionary lemma} 
					& Synonyms  & 65    & 34    & 45 \\
					& Hypernyms & 73    & 38    & 50 \\
					& Hyponyms  & 72    & 57    & 63 \\
					& Antonyms  & 58    & 71    & 48 \\
					& All       & 68    & 49    & 52 \\
					\bottomrule
				\end{tabularx}
				\normalsize
				\caption{\label{tab:SenseFitting-disambig} F$_1$- and precision (P) scores for relation disambiguation of SenseFitting in percent.}
			\end{table}
			
			\Fref{tab:SenseFitting-disambig} shows the F$_1$-scores of disambiguating these relations.
			The GermaNet disambiguation reaches a score of more than $90\%$ for all relation types with a precision of $94 - 100\%$.
			We assume that this high performance is due to GermaNet's structure: 
			all relations are symmetric\footnote{Each hyponym relation has a hypernym relation in reverse direction, so we consider it to be symmetric, too.}, and relations between synsets essentially create fully connected graphs of senses that all get moved towards their center.
			
			Wiktionary does not show the graph structure of GermaNet; nor is its development controlled by a group of experts, but by a heterogeneous online community based on the wiki principle.
			\citet{Mehler:Gleim:2018} show that this community consists of a small number of highly active authors that contribute to a broader set of topics and a large number of less active authors who contribute to a narrower set of topics.
			This imbalance suggests that Wiktionary's thematic orientation may be distorted by these highly active authors.
			As such, Wiktionary is likely not as consistent and stable as a resource created by a group of experts.
			
			The performance of Wiktionary's relation disambiguation on the dataset from \citet{Meyer:Gurevych:2012} confirms this hypothesis.
			Although precision remains high around 70\%, F$_1$-scores drop about half, because recall is rather low.
			Synonyms show the most significant difference to results obtained for GermaNet. 
			We assume this to be caused by the lack of symmetry in Wiktionary relations which is present in about 20\% of synonymy relations\footnote{For Wiktionary sense-to-lemma relations, we consider a relation to be symmetric if $\{(S_a, L_b), (S_a, L_b)\} \subseteq R_r$, where $S_X$ is a sense of lemma $L_X$ and $R_r$ is the set of all relations of the given type $r$.}.
			
			These results show that our semantic specialization can produce viable results for relation disambiguation if the used resource is well structured and features dense relations.
			
		\subsubsection{SenseFitting Training Resources}
		\label{sec:supplement-sensefitting-training-resources}
		
		    \Fref{tab:sense-inv-stats} shows the relation count for each resource in all variations.
		    The before mentioned graph structure of GermaNet can be seen in the count of hyponyms and hypernyms: they are equal as both relations are symmetric.
		    This symmetry helps drawing hypernym-hyponym pairs together, most likely increasing SenseFittings positive effect on injecting semantic information into sense embeddings.
		
    	    \begin{table}[ht!]
    			\centering
    			\small
    			\begin{tabularx}{\linewidth}{Xrrrr}
    				\toprule
    				Dataset & Syn & Hypo & Hyper & Ant \\
    				\midrule
    				GN & $103\,554$ & $314\,058$ & $314\,058$ & $3\,486$ \\ % statistics
    				GN$^+$ & $95\,740$ & $294\,872$ & $294\,872$ & $3\,426$ \\ % actually used during training.
    				\midrule
    				WKT & $93\,800$ & $182\,224$ & $124\,724$ & $82\,675$ \\ % statistics
    				WKT$^+$ & $50\,335$ & $114\,762$ & $85\,349$ & $51\,173$ \\ % actually used during training
    				lower$^*$ & $31\,352$ & $46\,053$ & $58\,002$ & $29\,314$ \\ % automatically disambiguated relations
    				lemma$^*$ & $29\,467$ & $43\,788$ & $53\,322$ & $28\,393$ \\
    				\bottomrule
    			\end{tabularx}
    			\normalsize
    			\caption{\label{tab:sense-inv-stats} The number of relations used in SenseFitting based on COW embeddings. 
    			$^+$: total number of applicable relations used in training; 
    			$^*$: number of automatically disambiguated relations for Wiktionary.}%\vspace*{-10pt}
    		\end{table}

	\subsection{Preprocessing WebCAGe}
	\label{sec:supplement-wca}
		
		\begin{table}[hbt!] % TODO: Table placement
			\centering
			\small
			\begin{tabularx}{\linewidth}{Xcccc}
				\toprule
				Task & Nouns & Verbs & Adjectives & Total \\
				\midrule
				WCA & 2\,150 & 1\,639 & 336 & 4\,125 \\  
				\midrule
				Senseval-2 & 1\,740 & 1\,806 & 375 & 3\,921 \\
				Senseval-3 & 1\,807 & 1\,978 & 159 & 3\,944 \\
				SemEval-07 & 2\,559 & 2\,292 & | & 4\,851 \\
				\bottomrule
			\end{tabularx}
			\normalsize
			\caption{\label{tab:WCA-stat} Number of sentences per part of speech in the WCA dataset in comparison to English datasets used for performing WSD in \citet{Iacobacci:Pilehvar:2016}.}
		\end{table}
		
		As mentioned in \fref{sec:supplement-wca}, WebCAGe is a corpus mainly composed of examples from word sense definitions from the German Wiktionary which were automatically annotated with GermaNet senses and hand-corrected afterward \citep{Henrich:Hinrichs:2011}.
		However, the authors do not provide a mapping to Wiktionary; only GermaNet senses are tagged with an ID.
    	Thus we needed to map WebCAGe back to Wiktionary to be able to evaluate the performance of our WSD method with both Wiktionary and GermaNet.
    	We were able to match a large portion of all sentences from WebCAGe to a recent German Wiktionary export from April 2018.
    	From this portion, we excluded sentences that did not include the target word itself (e.g.\ if a target word was only contained in the sentence as part of a word formation or compound)\footnote{For example, an excluded instance of the target word \textit{Bank} is \enquote{Bankleitzahl} (English: bank code), a compound of the words \textit{Bank} and \textit{Leitzahl}.}. % TODO: cite word formation/compound use in German?
		The size of WCA is similar to the size of datasets available for WSD tasks in English.
		\Fref{tab:WCA-stat} compares WCA against task-related datasets used to evaluate the method of \citet{Iacobacci:Pilehvar:2016}, namely Senseval-2 \citep{Edmonds:Cotton:2001}, Senseval-3 \citep{Mihalcea:Chklovski:2004} and SemEval-07 \citep{Pradhan:Loper:2007}. 
		Note that WCA differs from the English datasets in the number of sentences per lemma: 
		It has on average only 1.89 sentences per lemma, while Senseval-2 has 59.3. % sentences per lemma.
		This might affect a WSD performance evaluation, as the methods performance is not evaluated as consistently across many samples for the same target word, but rather sparsely across a large variety of target words.
		While WebCAGe (and WCA alike) might cover many different words, it does not cover all possible senses for a given target word.
		% TODO Review#2: Why is this information given? Add an interpretation or remove!

\end{document}

%% file: material_colors.tex
\definecolor{red50}{HTML}{ffebee}
\definecolor{red100}{HTML}{ffcdd2}
\definecolor{red200}{HTML}{ef9A9a}
\definecolor{red300}{HTML}{e57373}
\definecolor{red400}{HTML}{ef5350}
\definecolor{red500}{HTML}{f44336}
\definecolor{red600}{HTML}{e53935}
\definecolor{red700}{HTML}{d32f2f}
\definecolor{red800}{HTML}{c62828}
\definecolor{red900}{HTML}{b71c1c}
\definecolor{redA100}{HTML}{ff8A80}
\definecolor{redA200}{HTML}{ff5252}
\definecolor{redA400}{HTML}{ff1744}
\definecolor{redA700}{HTML}{d50000}
\definecolor{pink50}{HTML}{fce4ec}
\definecolor{pink100}{HTML}{f8bbd0}
\definecolor{pink200}{HTML}{f48fb1}
\definecolor{pink300}{HTML}{f06292}
\definecolor{pink400}{HTML}{ec407a}
\definecolor{pink500}{HTML}{e91e63}
\definecolor{pink600}{HTML}{d81b60}
\definecolor{pink700}{HTML}{c2185b}
\definecolor{pink800}{HTML}{ad1457}
\definecolor{pink900}{HTML}{880e4f}
\definecolor{pinkA100}{HTML}{ff80ab}
\definecolor{pinkA200}{HTML}{ff4081}
\definecolor{pinkA400}{HTML}{f50057}
\definecolor{pinkA700}{HTML}{c51162}
\definecolor{purple50}{HTML}{f3e5f5}
\definecolor{purple100}{HTML}{e1bee7}
\definecolor{purple200}{HTML}{ce93d8}
\definecolor{purple300}{HTML}{bA68c8}
\definecolor{purple400}{HTML}{ab47bc}
\definecolor{purple500}{HTML}{9c27b0}
\definecolor{purple600}{HTML}{8e24aa}
\definecolor{purple700}{HTML}{7b1fA2}
\definecolor{purple800}{HTML}{6A1b9a}
\definecolor{purple900}{HTML}{4A148c}
\definecolor{purpleA100}{HTML}{eA80fc}
\definecolor{purpleA200}{HTML}{e040fb}
\definecolor{purpleA400}{HTML}{d500f9}
\definecolor{purpleA700}{HTML}{aA00ff}
\definecolor{deeppurple50}{HTML}{ede7f6}
\definecolor{deeppurple100}{HTML}{d1c4e9}
\definecolor{deeppurple200}{HTML}{b39ddb}
\definecolor{deeppurple300}{HTML}{9575cd}
\definecolor{deeppurple400}{HTML}{7e57c2}
\definecolor{deeppurple500}{HTML}{673ab7}
\definecolor{deeppurple600}{HTML}{5e35b1}
\definecolor{deeppurple700}{HTML}{512dA8}
\definecolor{deeppurple800}{HTML}{4527A0}
\definecolor{deeppurple900}{HTML}{311b92}
\definecolor{deeppurpleA100}{HTML}{b388ff}
\definecolor{deeppurpleA200}{HTML}{7c4dff}
\definecolor{deeppurpleA400}{HTML}{651fff}
\definecolor{deeppurpleA700}{HTML}{6200ea}
\definecolor{indigo50}{HTML}{e8eaf6}
\definecolor{indigo100}{HTML}{c5cae9}
\definecolor{indigo200}{HTML}{9fA8da}
\definecolor{indigo300}{HTML}{7986cb}
\definecolor{indigo400}{HTML}{5c6bc0}
\definecolor{indigo500}{HTML}{3f51b5}
\definecolor{indigo600}{HTML}{3949ab}
\definecolor{indigo700}{HTML}{303f9f}
\definecolor{indigo800}{HTML}{283593}
\definecolor{indigo900}{HTML}{1A237e}
\definecolor{indigoA100}{HTML}{8c9eff}
\definecolor{indigoA200}{HTML}{536dfe}
\definecolor{indigoA400}{HTML}{3d5afe}
\definecolor{indigoA700}{HTML}{304ffe}
\definecolor{blue50}{HTML}{e3f2fd}
\definecolor{blue100}{HTML}{bbdefb}
\definecolor{blue200}{HTML}{90caf9}
\definecolor{blue300}{HTML}{64b5f6}
\definecolor{blue400}{HTML}{42A5f5}
\definecolor{blue500}{HTML}{2196f3}
\definecolor{blue600}{HTML}{1e88e5}
\definecolor{blue700}{HTML}{1976d2}
\definecolor{blue800}{HTML}{1565c0}
\definecolor{blue900}{HTML}{0d47A1}
\definecolor{blueA100}{HTML}{82b1ff}
\definecolor{blueA200}{HTML}{448aff}
\definecolor{blueA400}{HTML}{2979ff}
\definecolor{blueA700}{HTML}{2962ff}
\definecolor{lightblue50}{HTML}{e1f5fe}
\definecolor{lightblue100}{HTML}{b3e5fc}
\definecolor{lightblue200}{HTML}{81d4fa}
\definecolor{lightblue300}{HTML}{4fc3f7}
\definecolor{lightblue400}{HTML}{29b6f6}
\definecolor{lightblue500}{HTML}{03A9f4}
\definecolor{lightblue600}{HTML}{039be5}
\definecolor{lightblue700}{HTML}{0288d1}
\definecolor{lightblue800}{HTML}{0277bd}
\definecolor{lightblue900}{HTML}{01579b}
\definecolor{lightblueA100}{HTML}{80d8ff}
\definecolor{lightblueA200}{HTML}{40c4ff}
\definecolor{lightblueA400}{HTML}{00b0ff}
\definecolor{lightblueA700}{HTML}{0091ea}
\definecolor{cyan50}{HTML}{e0f7fa}
\definecolor{cyan100}{HTML}{b2ebf2}
\definecolor{cyan200}{HTML}{80deea}
\definecolor{cyan300}{HTML}{4dd0e1}
\definecolor{cyan400}{HTML}{26c6da}
\definecolor{cyan500}{HTML}{00bcd4}
\definecolor{cyan600}{HTML}{00acc1}
\definecolor{cyan700}{HTML}{0097A7}
\definecolor{cyan800}{HTML}{00838f}
\definecolor{cyan900}{HTML}{006064}
\definecolor{cyanA100}{HTML}{84ffff}
\definecolor{cyanA200}{HTML}{18ffff}
\definecolor{cyanA400}{HTML}{00e5ff}
\definecolor{cyanA700}{HTML}{00b8d4}
\definecolor{teal50}{HTML}{e0f2f1}
\definecolor{teal100}{HTML}{b2dfdb}
\definecolor{teal200}{HTML}{80cbc4}
\definecolor{teal300}{HTML}{4db6ac}
\definecolor{teal400}{HTML}{26A69a}
\definecolor{teal500}{HTML}{009688}
\definecolor{teal600}{HTML}{00897b}
\definecolor{teal700}{HTML}{00796b}
\definecolor{teal800}{HTML}{00695c}
\definecolor{teal900}{HTML}{004d40}
\definecolor{tealA100}{HTML}{A7ffeb}
\definecolor{tealA200}{HTML}{64ffda}
\definecolor{tealA400}{HTML}{1de9b6}
\definecolor{tealA700}{HTML}{00bfA5}
\definecolor{green50}{HTML}{e8f5e9}
\definecolor{green100}{HTML}{c8e6c9}
\definecolor{green200}{HTML}{A5d6A7}
\definecolor{green300}{HTML}{81c784}
\definecolor{green400}{HTML}{66bb6a}
\definecolor{green500}{HTML}{4caf50}
\definecolor{green600}{HTML}{43A047}
\definecolor{green700}{HTML}{388e3c}
\definecolor{green800}{HTML}{2e7d32}
\definecolor{green900}{HTML}{1b5e20}
\definecolor{greenA100}{HTML}{b9f6ca}
\definecolor{greenA200}{HTML}{69f0ae}
\definecolor{greenA400}{HTML}{00e676}
\definecolor{greenA700}{HTML}{00c853}
\definecolor{lightgreen50}{HTML}{f1f8e9}
\definecolor{lightgreen100}{HTML}{dcedc8}
\definecolor{lightgreen200}{HTML}{c5e1A5}
\definecolor{lightgreen300}{HTML}{aed581}
\definecolor{lightgreen400}{HTML}{9ccc65}
\definecolor{lightgreen500}{HTML}{8bc34a}
\definecolor{lightgreen600}{HTML}{7cb342}
\definecolor{lightgreen700}{HTML}{689f38}
\definecolor{lightgreen800}{HTML}{558b2f}
\definecolor{lightgreen900}{HTML}{33691e}
\definecolor{lightgreenA100}{HTML}{ccff90}
\definecolor{lightgreenA200}{HTML}{b2ff59}
\definecolor{lightgreenA400}{HTML}{76ff03}
\definecolor{lightgreenA700}{HTML}{64dd17}
\definecolor{lime50}{HTML}{f9fbe7}
\definecolor{lime100}{HTML}{f0f4c3}
\definecolor{lime200}{HTML}{e6ee9c}
\definecolor{lime300}{HTML}{dce775}
\definecolor{lime400}{HTML}{d4e157}
\definecolor{lime500}{HTML}{cddc39}
\definecolor{lime600}{HTML}{c0cA33}
\definecolor{lime700}{HTML}{afb42b}
\definecolor{lime800}{HTML}{9e9d24}
\definecolor{lime900}{HTML}{827717}
\definecolor{limeA100}{HTML}{f4ff81}
\definecolor{limeA200}{HTML}{eeff41}
\definecolor{limeA400}{HTML}{c6ff00}
\definecolor{limeA700}{HTML}{aeeA00}
\definecolor{yellow50}{HTML}{fffde7}
\definecolor{yellow100}{HTML}{fff9c4}
\definecolor{yellow200}{HTML}{fff59d}
\definecolor{yellow300}{HTML}{fff176}
\definecolor{yellow400}{HTML}{ffee58}
\definecolor{yellow500}{HTML}{ffeb3b}
\definecolor{yellow600}{HTML}{fdd835}
\definecolor{yellow700}{HTML}{fbc02d}
\definecolor{yellow800}{HTML}{f9A825}
\definecolor{yellow900}{HTML}{f57f17}
\definecolor{yellowA100}{HTML}{ffff8d}
\definecolor{yellowA200}{HTML}{ffff00}
\definecolor{yellowA400}{HTML}{ffeA00}
\definecolor{yellowA700}{HTML}{ffd600}
\definecolor{amber50}{HTML}{fff8e1}
\definecolor{amber100}{HTML}{ffecb3}
\definecolor{amber200}{HTML}{ffe082}
\definecolor{amber300}{HTML}{ffd54f}
\definecolor{amber400}{HTML}{ffcA28}
\definecolor{amber500}{HTML}{ffc107}
\definecolor{amber600}{HTML}{ffb300}
\definecolor{amber700}{HTML}{ffA000}
\definecolor{amber800}{HTML}{ff8f00}
\definecolor{amber900}{HTML}{ff6f00}
\definecolor{amberA100}{HTML}{ffe57f}
\definecolor{amberA200}{HTML}{ffd740}
\definecolor{amberA400}{HTML}{ffc400}
\definecolor{amberA700}{HTML}{ffab00}
\definecolor{orange50}{HTML}{fff3e0}
\definecolor{orange100}{HTML}{ffe0b2}
\definecolor{orange200}{HTML}{ffcc80}
\definecolor{orange300}{HTML}{ffb74d}
\definecolor{orange400}{HTML}{ffA726}
\definecolor{orange500}{HTML}{ff9800}
\definecolor{orange600}{HTML}{fb8c00}
\definecolor{orange700}{HTML}{f57c00}
\definecolor{orange800}{HTML}{ef6c00}
\definecolor{orange900}{HTML}{e65100}
\definecolor{orangeA100}{HTML}{ffd180}
\definecolor{orangeA200}{HTML}{ffab40}
\definecolor{orangeA400}{HTML}{ff9100}
\definecolor{orangeA700}{HTML}{ff6d00}
\definecolor{deeporange50}{HTML}{fbe9e7}
\definecolor{deeporange100}{HTML}{ffccbc}
\definecolor{deeporange200}{HTML}{ffab91}
\definecolor{deeporange300}{HTML}{ff8A65}
\definecolor{deeporange400}{HTML}{ff7043}
\definecolor{deeporange500}{HTML}{ff5722}
\definecolor{deeporange600}{HTML}{f4511e}
\definecolor{deeporange700}{HTML}{e64A19}
\definecolor{deeporange800}{HTML}{d84315}
\definecolor{deeporange900}{HTML}{bf360c}
\definecolor{deeporangeA100}{HTML}{ff9e80}
\definecolor{deeporangeA200}{HTML}{ff6e40}
\definecolor{deeporangeA400}{HTML}{ff3d00}
\definecolor{deeporangeA700}{HTML}{dd2c00}
\definecolor{brown50}{HTML}{efebe9}
\definecolor{brown100}{HTML}{d7ccc8}
\definecolor{brown200}{HTML}{bcaaA4}
\definecolor{brown300}{HTML}{A1887f}
\definecolor{brown400}{HTML}{8d6e63}
\definecolor{brown500}{HTML}{795548}
\definecolor{brown600}{HTML}{6d4c41}
\definecolor{brown700}{HTML}{5d4037}
\definecolor{brown800}{HTML}{4e342e}
\definecolor{brown900}{HTML}{3e2723}
\definecolor{gray50}{HTML}{fafafa}
\definecolor{gray100}{HTML}{f5f5f5}
\definecolor{gray200}{HTML}{eeeeee}
\definecolor{gray300}{HTML}{e0e0e0}
\definecolor{gray400}{HTML}{bdbdbd}
\definecolor{gray500}{HTML}{9e9e9e}
\definecolor{gray600}{HTML}{757575}
\definecolor{gray700}{HTML}{616161}
\definecolor{gray800}{HTML}{424242}
\definecolor{gray900}{HTML}{212121}
\definecolor{bluegray50}{HTML}{eceff1}
\definecolor{bluegray100}{HTML}{cfd8dc}
\definecolor{bluegray200}{HTML}{b0bec5}
\definecolor{bluegray300}{HTML}{90A4ae}
\definecolor{bluegray400}{HTML}{78909c}
\definecolor{bluegray500}{HTML}{607d8b}
\definecolor{bluegray600}{HTML}{546e7a}
\definecolor{bluegray700}{HTML}{455A64}
\definecolor{bluegray800}{HTML}{37474f}
\definecolor{bluegray900}{HTML}{263238}

%% file: colors.tex
\definecolorset{cmyk}{GU-}{}{%
  Goethe-Blau,1,0.2,0,0.4;%
  Hellgrau,0.04,0.04,0.05,0.02;%
  Sandgrau,0.12,0.09,0.13,0;%
  Dunkelgrau,0.25,0.25,0.3,0.75;%
  Purple,0.08,1,0.3,0.36;%
  Emo-Rot,0.04,1,0.8,0.07;%
  Senfgelb,0.01,0.25,1,0.05;%
  Gruen,0.62,0.4,0.87,0.09;%
  Magenta,0.08,0.86,0.12,0.12;%
  Orange,0,0.7,1,0.04;%
  Sonnengelb,0,0.12,0.95,0;%
  Helles-Gruen,0.4,0.17,0.81,0.07;%
  Lichtblau,0.8,0,0.06,0.04}
  
\definecolor{HUDarkBlue}{RGB}{16,82,132}
\definecolor{HULightBlue}{RGB}{55,140,195} %% HTML: 378cc3
\definecolor{HUBlueText}{RGB}{0,115,186}
\definecolor{HUDarkGrey}{RGB}{206,207,209}
\definecolor{HUMedGrey}{RGB}{219,220,221}
\definecolor{HULightGrey}{RGB}{241,241,241} %% HTML: f1f1f1
\definecolor{HUDarkOrange}{RGB}{146,76,0} %% HTML: 924c00
\definecolor{LNshadowblue}{RGB}{112,159,203}
\definecolor{LNlightblue}{RGB}{153,196,236}
\definecolor{LNultralightblue}{RGB}{229,240,250}
\definecolor{LNgreen}{RGB}{153,204,51}
\definecolor{GoetheOrange}{RGB}{243,192,57} % Pantone 143
\definecolor{GoetheOrangePlakativ}{RGB}{237,167,45} % Pantone 151
\definecolor{Papier}{RGB}{233,234,192}
\definecolor{agttlightSeminarGrau}{RGB}{219,220,222}
\definecolor{bottomcolor}{rgb}{0.32,0.3,0.38}
\definecolor{middlecolor}{rgb}{0.08,0.08,0.16}
\definecolor{loeweblau}{rgb}{0.33,0.57,0.84}
\definecolor{loewehellblau}{RGB}{219,238,244}
\definecolor{loewegruen}{RGB}{90,162,80}
\definecolor{loeweorange}{RGB}{220,150,55}
\definecolor{DigihumRot}{RGB}{212,0,0}
\definecolor{DigihumBlau}{RGB}{0,94,170}
\definecolor{DigihumHellBlau}{RGB}{221,240,255}
\definecolor{DigihumText}{RGB}{103,102,102}
\definecolor{DigihumHellGrau}{RGB}{214,214,214}
\colorlet{loewegrau}{DigihumText!83!white}
\definecolor{SeminarBlau}{RGB}{0,154,224}
\definecolor{SeminarRot}{rgb}{0.75,0,0}
\definecolor{SeminarGruen}{RGB}{0,128,0}
\definecolor{SeminarOrange}{RGB}{237,167,45} % rgb: {0.93,0.65,0.18}
\definecolor{SeminarGrau}{rgb}{0.32,0.3,0.38} % RGB: {81.6,76.5,96.6}
\colorlet{SeminarHellBlau}{SeminarBlau!25!white}
\colorlet{SeminarHellGruen}{SeminarGruen!25!white}
\colorlet{SeminarHellRot}{SeminarRot!25!white}
\colorlet{SeminarHellOrange}{GoetheOrangePlakativ!25!white}
\colorlet{SeminarHellGrau}{SeminarGrau!35!white}
\colorlet{SeminarSehrHellGrau}{SeminarGrau!15!white}
\colorlet{SeminarSehrSehrHellGrau}{SeminarGrau!5!white}

\colorlet{SeminarBlauGruen}{SeminarBlau!50!SeminarGruen}
\colorlet{SeminarGruenBlau}{SeminarBlau!50!SeminarGruen}
\colorlet{SeminarMix}{SeminarBlau!50!SeminarGruen}
\colorlet{SeminarHellMix}{SeminarMix!25!white}
\colorlet{SeminarHellBlauGruen}{SeminarMix!25!white}
\colorlet{SeminarHellGruenBlau}{SeminarMix!25!white}
\colorlet{SeminarRotOrange}{SeminarRot!50!GoetheOrangePlakativ}
\colorlet{SeminarOrangeRot}{SeminarRot!50!GoetheOrangePlakativ}
\colorlet{SeminarHellOrangeRot}{SeminarOrangeRot!25!white}
\colorlet{SeminarHellRotOrange}{SeminarOrangeRot!25!white}
\colorlet{SeminarRotGruen}{SeminarRot!50!SeminarGruen}
\colorlet{SeminarGruenRot}{SeminarRot!50!SeminarGruen}
\colorlet{SeminarRotBlau}{SeminarRot!50!SeminarBlau}
\colorlet{SeminarBlauRot}{SeminarRot!50!SeminarBlau}
\colorlet{SeminarHellRotBlau}{SeminarRotBlau!25!white}
\colorlet{SeminarHellBlauRot}{SeminarRotBlau!25!white}
\colorlet{SeminarHellRotGruen}{SeminarRotGruen!25!white}
\colorlet{SeminarHellGruenRot}{SeminarRotGruen!25!white}
\colorlet{SeminarOrangeGruen}{SeminarOrange!50!SeminarGruen}
\colorlet{SeminarGruenOrange}{SeminarOrange!50!SeminarGruen}
\colorlet{SeminarHellGruenOrange}{SeminarOrangeGruen!25!white}
\colorlet{SeminarHellOrangeGruen}{SeminarOrangeGruen!25!white}
\colorlet{SeminarOrangeBlau}{SeminarOrange!50!SeminarBlau}
\colorlet{SeminarBlauOrange}{SeminarOrange!50!SeminarBlau}
\colorlet{SeminarHellBlauOrange}{SeminarOrangeBlau!25!white}
\colorlet{SeminarHellOrangeBlau}{SeminarOrangeBlau!25!white}
\colorlet{SeminarSehrHellRot}{SeminarRot!15!white}
\colorlet{SeminarSehrHellGrau}{SeminarGrau!15!white}
\colorlet{SeminarSehrHellGruen}{SeminarGruen!15!white}
\colorlet{SeminarSehrHellBlau}{SeminarBlau!15!white}
\colorlet{SeminarSehrHellOrange}{SeminarHellOrange!15!white}
\colorlet{SeminarOrangeGrau}{SeminarOrange!50!SeminarGrau}
\colorlet{SeminarGrauOrange}{SeminarOrange!50!SeminarGrau}
\colorlet{SeminarRotGrau}{SeminarRot!50!SeminarGrau}
\colorlet{SeminarGrauRot}{SeminarRot!50!SeminarGrau}
\colorlet{SeminarBlauGrau}{SeminarBlau!50!SeminarGrau}
\colorlet{SeminarGrauBlau}{SeminarBlau!50!SeminarGrau}
\colorlet{SeminarGruenGrau}{SeminarGruen!50!SeminarGrau}
\colorlet{SeminarGrauGruen}{SeminarGruen!50!SeminarGrau}
\colorlet{SeminarHellGrauRot}{SeminarGrauRot!25!white}
\colorlet{SeminarHellRotGrau}{SeminarGrauRot!25!white}
\colorlet{SeminarHellGrauBlau}{SeminarGrauBlau!25!white}
\colorlet{SeminarHellBlauGrau}{SeminarGrauBlau!25!white}
\colorlet{SeminarHellGrauGruen}{SeminarGrauGruen!25!white}
\colorlet{SeminarHellGruenGrau}{SeminarGrauGruen!25!white}
\colorlet{SeminarHellGrauOrange}{SeminarGrauOrange!25!white}
\colorlet{SeminarHellOrangeGrau}{SeminarGrauOrange!25!white} 

%% file: tikz.tex
\usepackage{tikz} % Grafiken
\usetikzlibrary{babel}
\usetikzlibrary{arrows}
\usetikzlibrary{arrows.meta}
\usetikzlibrary{automata}
\usetikzlibrary{positioning}
\usetikzlibrary{matrix}
\usetikzlibrary{fit}
\usetikzlibrary{backgrounds}
\usetikzlibrary{shadows}
\usetikzlibrary{shapes.multipart}
\usetikzlibrary{shapes.misc}
\usetikzlibrary{shapes.geometric}
\usetikzlibrary{shapes.arrows}
\usetikzlibrary{shapes.symbols}
\usetikzlibrary{shapes.callouts}
\usetikzlibrary{er}
\usetikzlibrary{shadings}
\usetikzlibrary{topaths}
\usetikzlibrary{chains}
\usetikzlibrary{decorations.pathreplacing}
\usetikzlibrary{decorations.pathmorphing}
\usetikzlibrary{patterns}
\usetikzlibrary{calc}
\usetikzlibrary{mindmap}
\usetikzlibrary{through}

\usepackage{forest}

\usepackage{pgfplots}
\pgfplotsset{compat=newest}

\tikzset{obj/.style={draw=SeminarGrau, double=white, anchor=text, rectangle split,
    rectangle split parts=4, rectangle split part align={left},
    rectangle split part fill={SeminarBlau, SeminarGruen, SeminarGrau, SeminarRot},
    text=white},
  ent/.style={draw=SeminarGrau, double=white, anchor=text, rectangle split,
    rectangle split parts=3, rectangle split part align={center},
    rectangle split part fill={SeminarBlau, SeminarGruen, SeminarGrau},
    text=white},
  overtag/.style={font=\normalsize,fill=white},
  tag/.style={font=\normalsize},
  kuller/.style={-o},
  dreibein/.style={-angle 90 reversed, shorten >=1pt},
  kullerdreibein/.style={o-angle 90 reversed, shorten >=1pt},
  dreibeinB/.style={angle 90 reversed-angle 90 reversed, shorten
    >=1pt, shorten <=1pt},
  kreis/.style={circle, draw, fill=white, inner sep=0pt, minimum
    width=4pt},
  association/.style={ellipse split, fill=LNlightblue, draw=SeminarGrau,
    double=white, drop shadow, text=black, inner sep=2pt},
  isa/.style={semicircle, fill=LNlightblue},
  phenomenon/.style={draw=SeminarGrau, double=white, drop shadow,
    fill=LNshadowblue, text=white, align=center, minimum height=1.6cm,
    text width=3.3cm},
  modell/.style={draw=SeminarGrau, double=white, drop shadow,
    fill=SeminarGruen, text=white, align=center, minimum height=1.6cm,
    text width=3.3cm},
  formalism/.style={draw=SeminarGrau, double=white, drop shadow,
    fill=black, text=white, align=center, minimum height=1.6cm,
    text width=3.3cm},
  instanziiert/.style={stealth-, font=\footnotesize, thick}
  eLexTable/.style={draw=SeminarGrau, double=white, anchor=text, rectangle split,
    rectangle split parts=3, rectangle split part align={center},
    rectangle split part fill={SeminarGrau!40, LNlightblue, white},
    text=black},
  some/.style={circle, draw=SeminarGrau, double=white, drop shadow,
    fill=LNshadowblue, text=white, align=center},
  korpus/.style={draw=SeminarGrau, double=white, fill=SeminarGruen,
    text=white, tape, tape bend top=none, text width=1.4cm, minimum
    height=2.1cm, align=center, double copy
    shadow={draw=SeminarGrau, thick, opacity=0.6, shadow xshift=1ex, shadow
      yshift=1ex}},
  aufbereitung/.style={rectangle, draw=SeminarGrau, double=white, drop shadow,
    fill=#1, text=white, text width=2.3cm, minimum
    height=1.2cm, align=center},
  schnittstelle/.style={circle, draw=SeminarGrau, fill=SeminarGrau,
    double=white, drop shadow, minimum size=1cm},
  instanz/.style={fill=SeminarRot,text=white, rectangle, thick,
    draw=SeminarGrau, font=\footnotesize},
  modell/.style={chamfered rectangle, draw=SeminarGrau, double=white, drop
    shadow, text=white, chamfered rectangle sep=0.2cm,
    fill=SeminarBlau, chamfered rectangle corners=north west, text
    width=1.8cm, minimum height=2.3cm, align=flush left},
  dokument/.style={cylinder, shape border rotate=90, aspect=0.25,
    fill=SeminarBlau, draw=SeminarGrau, double=white, text=white, text
    width=1.4cm, align=center, drop shadow,
    minimum height=2.3cm},
  eLex/.style={loose background, top color=SeminarGruen!40,
    bottom color=LNlightblue, middle color=SeminarGrau!20, % rounded corners,
    draw=SeminarGrau},
  defnode/.style={rectangle, rounded corners, drop shadow,
    fill=LNultralightblue, text width=11cm, align=flush left},
  zitterkreis/.style={draw=black, decorate, decoration={random steps,
    segment length=3pt, amplitude=0.5pt}, circle, line width=1.5pt,
    minimum size=1.1cm},
  rectext/.style={rectangle, font=\TTBOLD, fill=white, align=center},
  onlytext/.style={font=\TTBOLD, fill=white, align=center},
  zitterquadrat/.style={rectangle, draw=black, line width=1.5pt,
    decorate, decoration={random steps, segment length=3pt,
    amplitude=0.4pt}, minimum size=1.3cm},
  zitterlinie/.style={draw, thick, decoration={random steps, segment
    length=15pt, amplitude=0.75pt}},
rectangular vjoint/.style={to path={-- ++(0,-0.6) -| (\tikztotarget)}},
rectangular hjoint/.style={to path={-- ++(-0.6,0) -| (\tikztotarget)}},
Hinweis/.style={circle, fill=SeminarRot, align=center, text width=2cm, text=white}
}

%% file: wiktionary_example.tikz
\begin{tikzpicture}[Bullet/.style={circle, inner sep=0pt, minimum size=3pt, fill=black}]

\matrix [draw, matrix of nodes, row sep=1pt, nodes={anchor=west}] (Senses links) {\textbf{Senses} \\ \enspace {[1]} Sense 1 \\ \enspace {[2]} Sense 2 \\ \enspace {[3]} Sense 3 \\};
\node [rectangle, text width=2.02cm, align=left, above=2pt of Senses links, fill=SeminarHellBlau] (LemmaEntry links) {Lemma};
\node [fit=(Senses links) (LemmaEntry links), draw] (Lemma links) {};
%  \node [rectangle, text width=2.28cm, above=2pt of Lemma links, fill=SeminarSehrHellGrau] (SuperLemmaEntry links) {Superlemma};
%  \node [fit=(Lemma links) (SuperLemmaEntry links), draw] (SuperLemma links) {};

\begin{scope}[xshift=4cm]
\matrix [draw, matrix of nodes, row sep=1pt, nodes={anchor=west}] (Senses rechts) {\textbf{Senses} \\ \enspace {[1]} Sense 1 \\ \enspace {[2]} Sense 2 \\ \enspace {[3]} Sense 3 \\};
\node [rectangle, text width=2.02cm, align=left, above=2pt of Senses rechts, fill=SeminarHellBlau] (LemmaEntry rechts) {Lemma};
\node [fit=(Senses rechts) (LemmaEntry rechts), draw] (Lemma rechts) {};
%  \node [rectangle, text width=2.28cm, above=2pt of Lemma rechts, fill=SeminarSehrHellGrau] (SuperLemmaEntry rechts) {Superlemma};
%  \node [fit=(Lemma rechts) (SuperLemmaEntry rechts), draw] (SuperLemma rechts) {};
\end{scope}

\begin{scope}[->, thick]
\draw [dashed] (Senses links-2-1.east) to [bend left] ([xshift=1ex]Senses rechts-2-1.west);
\path [SeminarGruen] (Senses links-2-1.east) edge[bend left] node [above=3pt] {\scriptsize URL} (LemmaEntry rechts.west);
% \draw [SeminarGruen] (Senses links-2-1.east) to [bend left] (LemmaEntry rechts);
%  \draw [SeminarBlau] (LemmaEntry links.east) to [bend left] (LemmaEntry rechts.west);
%  \draw [SeminarGruen] (Senses links-2-1.east) to [bend left] (SuperLemmaEntry rechts);
%  \draw [SeminarBlau] (LemmaEntry links.east) to [bend left] (SuperLemmaEntry rechts.west);
\end{scope}

\end{tikzpicture}

%% file: inter-annotator_agreement.tikz
% Data	rho	1/sigma	sigma
% all	0.846	1.489	0.671
% positive	0.791749636	1.48427673	0.673728814
% negative 	0.666794212	1.571428571	0.636363636
% false	0.608	1.396	0.716
% has_antonym	0.605417469	2.133333333	0.46875
% has_hypernym	0.609047618	1.24137931	0.805555556
% has_hyponym	0.685992993	1.333333333	0.75
% has_synonym	0.509917842	1.7	0.588235294
\begin{tikzpicture}
    \begin{axis}[
        width=0.8\linewidth,
        height=2.5cm,
	    enlargelimits=0.25,
        scale only axis,
        ytick={0.6,0.7,0.8},
		minor tick num=1,
        xtick={1,2,3,4},
        x tick style={draw=none},
        x tick label style={yshift=-2pt,rotate=45,anchor=east},
        xticklabels={all, positive, negative, false},
        legend style={legend columns=-1},
        ybar=3pt,
	    nodes near coords,
        every node near coord/.append style={font=\tiny},
        bar width=14pt,
      ]
      
      \addplot[red800, fill] coordinates {(1,0.846) (2,0.791749636) (3,0.666794212) (4,0.608)};
      \addplot[blue500, fill] coordinates {(1,0.671) (2,0.673728814) (3,0.636363636) (4,0.716)};
      
      \legend{$\rho$, $\sigma$}
    \end{axis}
    % \begin{axis}[
    %     width=0.8\linewidth,
    %     height=3cm,
    %     scale only axis,
    %     clip=false,
    %     separate axis lines,
    %     axis on top,
    %     ymin=0.5, ymax=0.9,
    %     ytick={0.6,0.7,0.8},
    %     xmin=0, xmax=5,
    %     xtick={1,2,3,4},
    %     x tick style={draw=none},
    %     x tick label style={xshift=4pt,yshift=-4pt,rotate=45,anchor=east},
    %     xticklabels={all, positive, negative, false},
    %     % ylabel={Avg. Response Deviation $\sigma$},
    %     ylabel={$\sigma$},
    %     ylabel style={rotate=-90, at={(-0.025,1)}},
    %     every axis plot/.append style={
    %       ybar,
    %       bar width=.75,
    %       bar shift=0pt,
    %       fill
    %     }
    %   ]
    %   \addplot[redA700]coordinates {(1,0.683333333)};
    %   \addplot[green600]coordinates{(2,0.673728814)};
    %   \addplot[purple800]coordinates{(3,0.636363636)};
    %   \addplot[orange400]coordinates{(4,0.783783784)};
    % \end{axis}
\end{tikzpicture}